\documentclass{article}
\usepackage{spconf,amsmath,epsfig,graphicx}
\usepackage{multirow}
\usepackage{url}
\begin{document}\sloppy
\def\x{{\mathbf x}}
\def\L{{\cal L}}

\title{Analyzing the group sparsity based on the rank minimization methods}
%
\name{\small{Zhiyuan Zha$^{1}$,  Xin Liu$^{2}$, Xiaohua Huang $^{2}$, Henglin Shi $^{2}$, Yingyue Xu $^{2}$, Qiong Wang $^{1,\star}$, Lan Tang $^{1}$, Xinggan Zhang $^{1}$}
\thanks{${\star}$ Corresponding Author.  }  }

\address{$^{1}$ \footnotesize{School of Electronic Science and Engineering, Nanjing University, Nanjing 210023, China.} \\
    $^{2}$ \footnotesize{The Center for Machine Vision and Signal Analysis, University of Oulu, 90014, Finland.}\\
}

\maketitle

\begin{abstract}
Sparse coding has achieved a great success in various image processing studies. However, there is not any benchmark to measure the sparsity of image patch/group because sparse discriminant conditions cannot keep unchanged. This paper analyzes the sparsity of group based on the strategy of the rank minimization. Firstly, an adaptive dictionary for each group is designed. Then, we prove that group-based sparse coding is equivalent to the rank minimization problem, and thus the sparse coefficient of each group are measured by estimating the singular values of each group. Based on that measurement, the weighted Schatten $p$-norm minimization (WSNM) has been found to be the closest solution to the real singular values of each group. Thus, WSNM can be equivalently transformed into a non-convex $\ell_p$-norm minimization problem in group-based sparse coding. To make the proposed scheme tractable and robust, the alternating direction method of multipliers (ADMM) is used to solve the  $\ell_p$-norm minimization problem. Experimental results on two applications: image inpainting and image compressive sensing (CS) recovery have shown that the proposed scheme outperforms many state-of-the-art methods.
\end{abstract}
\section{Introduction}
\label{sec:intro}

Traditional patch-based sparse coding assumes that each patch of an image can be precisely modeled as a sparse linear combination of basic elements. It has been successfully used in various image processing applications \cite{1,2,3,4}. However, patch-based sparse coding model of natural images usually suffers from some limits, such as dictionary learning with great computational complexity, neglecting the relationship between similar patches.

Instead of using patch as the basic unit of sparse coding, group-based sparse coding can offer a powerful mechanism of combining local sparsity and nonlocal self-similarity of images simultaneously in a unified framework \cite{3,8}. To be concrete,  an image $\textbf{\emph{X}}$ with size $N$ is divided into $n$ overlapped patches of size $\sqrt{d}\times \sqrt{d}$, and each patch is denoted by the vector ${\textbf{\emph{x}}}_i\in\Re^d, i=1,2,...,n$. Then for each patch ${\textbf{\emph{x}}}_i$, its $k$ similar patches are selected from a $I \times I$ sized search window to form a set ${\textbf{\emph{S}}}_i$. After this, all the patches in ${\textbf{\emph{S}}}_i$ are stacked into a matrix ${\textbf{\emph{X}}}_i\in\Re^{d\times k}$, i.e., ${\textbf{\emph{X}}}_i=\{{\textbf{\emph{x}}}_{i,1}, {\textbf{\emph{x}}}_{i,2},...,{\textbf{\emph{x}}}_{i,k}\}$.  The matrix ${\textbf{\emph{X}}}_i$ consists of all the patches with similar structures is called  a group,  where ${\textbf{\emph{x}}}_{i,k}$ denotes the $k$-th similar patch of the $i$-th group.  Similar to patch-based sparse coding \cite{1}, given a dictionary ${\textbf{\emph{D}}}_{i}$, each group ${\textbf{\emph{X}}}_i$ can be sparsely represented as ${\boldsymbol\alpha}_i={\textbf{\emph{D}}}_i^{-1}{\textbf{\emph{X}}}_i$ and solved by the following $\ell_0$-norm minimization problem,
\begin{equation}
{\boldsymbol\alpha}_i=\arg\min_{{\boldsymbol\alpha}_i} \{\frac{1}{2}||{\textbf{\emph{X}}}_i-{\textbf{\emph{D}}}_i{\boldsymbol\alpha}_i||_F^2+\lambda||{\boldsymbol\alpha}_i||_0\}
\label{eq:1}
\end{equation} 
where $\lambda$ is the regularization parameter, $||*||_F^2$ denotes the Frobenius norm, and $||*||_0$ is $\ell_0$-norm, counting the non-zero entries of ${\boldsymbol\alpha}_i$.

However, since $||*||_0$ norm minimization is a difficult combinatorial optimization problem, solving  Eq.~\eqref{eq:1} is  NP-hard. For this reason, it is often replaced by the $\ell_1$-norm or the reweighted $\ell_1$-norm to make the optimization problem easy \cite{5}. Nonetheless, the solution of these norm minimizations is only the estimation of the real sparsity solution under certain conditions. For instance, Cand{\`e}s $\emph{et al}.$ \cite{10} proposed that solving $\ell_1$-norm optimization problem can recover a $K$-sparse signal $\textbf{\emph{x}}\in\Re^N$ from $M=O(Klog(N/K))$ random measurements in compressive sensing (CS). However, the  $\ell_1$-norm minimization cannot still obtain the real sparsity solution, one important reason is nobody can guarantee the invariance of sparse discriminant conditions. In other words, there is not any benchmark to measure the sparsity of a signal.

With the above consideration, we analyze the group sparsity from the point of the rank minimization. To the best of our knowledge, few works have exploited the rank minimization methods to analyze the sparsity of image groups. The contribution of this paper is as follows. First, an adaptive dictionary for each group is designed with a low computational complexity, rather than dictionary learning from natural images. Second, based on this dictionary learning scheme, we prove the equivalence of  group-based sparse coding and the rank minimization problem, and thus the sparse coefficient of each group  are measured by calculating the singular values of each group. Thus, we have a benchmark to measure the sparsity of each group because the singular values of the original image group can be easily computed by SVD operator. Third, we exploit four nuclear norms (i.e., standard nuclear norm \cite{11},  the weighted nuclear norm \cite{12}, Schatten $p$-norm \cite{13} and the weighted Schatten $p$-norm \cite{14}) to analyze the sparsity of each group and the solution of the weighted Schatten $p$-norm minimization (WSNM) is the nearest to real singular values of each group. Therefore, WSNM is equivalently turned into a non-convex $\ell_p$-norm minimization problem in group-based sparse coding. To make the proposed scheme tractable and robust, we exploit the alternating direction method of multipliers (ADMM) to solve the non-convex $\ell_p$-norm minimization problem.  Experimental results on two low-level vision tasks, i.e., image inpainting and image compressive sensing (CS) recovery have demonstrated that the proposed scheme outperforms many state-of-the-art schemes.

\section{Background}
\subsection{Rank minimization method}

The main goal of low rank matrix approximation (LRMA) is to recover the underlying low rank structure from its degraded/corrupted observed version. In general, methods of LRMA can be classified into two categories: the low rank matrix factorization (LRMF) methods \cite{15,16} and the rank minimization methods \cite{11,12,13,14}. In this work we focus on the latter category. More specifically, given an input matrix  $\textbf{\emph{Y}}$, the rank minimization methods aim to find  a low rank matrix $\textbf{\emph{X}}$, which is as close to $\textbf{\emph{Y}}$ as possible under $F$-norm data fidelity and one nuclear norm,
\begin{equation}
\hat{\textbf{\emph{X}}}={\arg\min}_\textbf{\emph{X}}||\textbf{\emph{Y}}-\textbf{\emph{X}}||_F^2 +\lambda \textbf{\emph{R}}(\textbf{\emph{X}})
\label{eq:2}
\end{equation}
where $\lambda$ is a trade-off parameter between the loss function and the low rank regularization induced by one nuclear norm $\textbf{\emph{R}}(\textbf{\emph{X}})$. We will briefly introduce several nuclear norms including standard nuclear norm \cite{11}, the weighted nuclear norm \cite{12}, Scappten $p$-norm \cite{13} and the weighted Scappten $p$-norm \cite{14} in the next subsection.

\subsection{Nuclear norms}
In this subsection, we first introduce the weighted Schatten $p$-norm \cite{14} of a matrix $\textbf{\emph{X}}\in\Re^{m\times n}$, which is defined as
\begin{equation}
||\textbf{\emph{X}}||_{\textbf{\emph{w}},S_p}=\left(\sum\nolimits_{i=1}^{min\{m,n\}}w_i\sigma_i^p\right)^{\frac{1}{p}}
\label{eq:3}
\end{equation}
where $0<p\leq 1$, and $\sigma_i$ is the $i$-th singular value of $\textbf{\emph{X}}$. $\textbf{\emph{w}}=[w_1,...,w_{min\{m,n\}}]$  and $w_i\geq0$ is a non-negative weight assigned to $\sigma_i$. Then the weighted Schatten $p$-norm of $\textbf{\emph{X}}$ with power $p$ is
\begin{equation}
||\textbf{\emph{X}}||_{\textbf{\emph{w}},S_p}^p=\left(\sum\nolimits_{i=1}^{min\{m,n\}}w_i\sigma_i^p \right)=Tr(\textbf{\emph{W}}{\boldsymbol\Delta}^p)
\label{eq:4}
\end{equation}
where $\textbf{\emph{W}}$ and ${\boldsymbol\Delta}$ are diagonal matrices whose diagonal entries are composed of $w_i$ and $\sigma_i$, respectively.

The Schatten $p$-norm \cite{13} of a matrix  $\textbf{\emph{X}}$ can be represented by setting $\textbf{\emph{w}}=[1,1,...,1]$ in Eq.~\eqref{eq:3},
\begin{equation}
||\textbf{\emph{X}}||_{S_p}=\left(\sum\nolimits_{i=1}^{min\{m,n\}}\sigma_i^p \right)^{\frac{1}{p}}=\left(Tr(({\textbf{\emph{X}}}^T{\textbf{\emph{X}}})^{\frac{p}{2}})\right)^{\frac{1}{p}}
\label{eq:5}
\end{equation}

The weighted nuclear norm \cite{12} of a matrix $\textbf{\emph{X}}$ can be represented by setting $p$=1 in Eq.~\eqref{eq:3},
\begin{equation}
||\textbf{\emph{X}}||_{\textbf{\emph{w},*}}=\left(\sum\nolimits_{i=1}^{min\{m,n\}}w_i\sigma_i \right)=Tr(\textbf{\emph{W}}{\boldsymbol\Delta})
\label{eq:6}
\end{equation}

A widely used standard nuclear norm \cite{11} of a matrix $\textbf{\emph{X}}$ can be represented by setting $p$=1 and $\textbf{\emph{w}}=[1,1,...,1]$ in Eq.~\eqref{eq:3},
\begin{equation}
||\textbf{\emph{X}}||_{*}=\sum\nolimits_{i=1}^{min\{m,n\}}\sigma_i=Tr(({\textbf{\emph{X}}}^T{\textbf{\emph{X}}})^{\frac{1}{2}})
\label{eq:7}
\end{equation}

\section{Analyzing the sparsity of group based on the rank minimization methods}
Since the sparse discriminant conditions cannot keep unchanged, there is not any benchmark to measure the sparsity of image group. Therefore, we analyze the group sparsity from the point of the rank minimization.  To this end, an adaptive dictionary for each group is designed with a low computational complexity, rather than dictionary learning from natural images. Based on this dictionary learning scheme, we prove the equivalence of group-based sparse coding and the rank minimization problem, i.e., the sparse coefficient of each group are measured by calculating the singular values of each group. Therefore, we possess a benchmark to measure the sparsity of each group by rank minimization methods since the singular values of the original image group can be easily obtained. In this way, we can achieve a clear visual comparison effect to analyze the sparsity of each group based on the rank minimization methods (See Fig.~\ref{fig:2}).
\subsection{Adaptive dictionary learning}
In this subsection, an adaptive dictionary learning method is designed, that is, for each group ${\textbf{\emph{X}}}_i$, its adaptive dictionary can be learned from its observation ${\textbf{\emph{Y}}}_i\in\Re^{d \times k}$.

More specifically, we apply the singular value decomposition (SVD) to ${\textbf{\emph{Y}}}_i$,
\begin{equation}
{\textbf{\emph{Y}}}_i= {\textbf{\emph{U}}}_i{\boldsymbol\Sigma}_i{\textbf{\emph{V}}}_i^T=\sum\nolimits_{j=1}^m \boldsymbol\sigma_{i,j}{\textbf{\emph{u}}}_{i,j}{\textbf{\emph{v}}}_{i,j}^T
\label{eq:8}
\end{equation}
where $\boldsymbol\mu_i=[\boldsymbol\sigma_{i,1},\boldsymbol\sigma_{i,2},...,\boldsymbol\sigma_{i,m}]$, $m={\rm min}(d,k)$, ${\boldsymbol\Sigma}_i={\rm diag}(\boldsymbol\mu_i)$ is a diagonal matrix whose non-zero elements are represented by $\boldsymbol\mu_i$,  and  ${\textbf{\emph{u}}}_{i,j}, {\textbf{\emph{v}}}_{i,j}$ are the columns of ${\textbf{\emph{U}}}_i$ and ${\textbf{\emph{V}}}_i$, respectively.

Moreover, we define each dictionary atom $\textbf{\emph{d}}_{i,j}$ of the adaptive dictionary $\textbf{\emph{D}}_i$ for each group $\textbf{\emph{Y}}_i$ as follows:
\begin{equation}
\textbf{\emph{d}}_{i,j}={\textbf{\emph{u}}}_{i,j}{\textbf{\emph{v}}}_{i,j}^T, \ \ \ j=1,2,...,m
\label{eq:9}
\end{equation}

Finally,  by learning an adaptive dictionary $\textbf{\emph{D}}_i=[\textbf{\emph{d}}_{i,1},\textbf{\emph{d}}_{i,2},...,\textbf{\emph{d}}_{i,m}]$ from each group ${\textbf{\emph{Y}}}_i$. The proposed dictionary learning method is efficient due to the fact that it only requires one SVD operator for each group.
\subsection{Prove the equivalence of group-based sparse coding and the rank minimization problem}
To prove that the group-based sparse coding is equivalent to the rank minimization problem, we firstly give following two lemmas.

\noindent$\textbf{{Lemma 1}}$ \ \ The minimization problem
\begin{equation}
{\textbf{\emph{x}}}=\arg\min_{\textbf{\emph{x}}} \frac{1}{2}||{\textbf{\emph{x}}}-{\textbf{\emph{a}}}||_2^2+\tau.||{\textbf{\emph{x}}}||_1
\label{eq:10}
\end{equation} 
has a closed form, which can be expressed as
\begin{equation}
\hat{\textbf{\emph{x}}}={\rm soft}({\textbf{\emph{a}}},\tau)= {\rm sgn}({\textbf{\emph{a}}},\tau). {\rm max}(abs({\textbf{\emph{a}}})-\tau,0)
\label{eq:11}
\end{equation} 

\emph{Proof:} see \cite{17}.

Consider the SVD of a matrix $\textbf{\emph{P}}\in\Re^{n_1\times n_2}$ of rank $r$
\begin{equation}
\textbf{\emph{P}}= \textbf{\emph{U}}\boldsymbol\Sigma\textbf{\emph{V}}^T, \boldsymbol\Sigma ={\rm diag}(\{\sigma_i\}_{1\leq i\leq r})
\label{eq:12}
\end{equation} 
where $\textbf{\emph{U}}\in\Re^{n_1 \times r}$ and $\textbf{\emph{V}}\in\Re^{n_2 \times r}$ are orthogonal matrices, respectively. $\sigma_i$ is the $i$-th singular value of $\textbf{\emph{P}}$. For each $\tau\geq0$, the soft-thresholding operator $\mathcal{D}_\tau$ is defined as
\begin{equation}
\mathcal{D}_\tau(\textbf{\emph{P}})= \textbf{\emph{U}} \mathcal{D}_\tau(\boldsymbol\Sigma)\textbf{\emph{V}}^T,\ \ \ \mathcal{D}_\tau(\boldsymbol\Sigma)= {\rm soft}(\sigma_i,\tau)
\label{eq:13}
\end{equation} 
\noindent$\textbf{{Lemma 2}}$ \ \ For each $\tau\geq0$, and $\textbf{\emph{Q}}\in\Re^{n_1\times n_2}$ ,the singular value shrinkage operator Eq.~\eqref{eq:13} satisfies
\begin{equation}
\mathcal{D}_\tau(\textbf{\emph{Q}})= \arg\min_{\textbf{\emph{P}}} \{\frac{1}{2}||{\textbf{\emph{P}}}-{\textbf{\emph{Q}}}||_F^2+\tau||{\textbf{\emph{P}}}||_*\}
\label{eq:14}
\end{equation} 

\emph{Proof:} see Appendix.

Now, the classical $\ell_1$-norm group-based sparse coding problem can be represented as
\begin{equation}
{\boldsymbol\alpha}_i=\arg\min\nolimits_{{\boldsymbol\alpha}_i} \{||{\textbf{\emph{Y}}}_i-{\textbf{\emph{D}}}_i{\boldsymbol\alpha}_i||_F^2+\lambda||{\boldsymbol\alpha}_i||_1\}
\label{eq:15}
\end{equation} 

According to the above design of adaptive dictionary ${\textbf{\emph{D}}}_i$, we have the following conclusion.

\noindent$\textbf{{Theorem 1}}$
\begin{equation}
||{\textbf{\emph{Y}}}_i-{\textbf{\emph{X}}}_i||_F^2=||\boldsymbol\mu_i-\boldsymbol\alpha_i||_2^2
\label{eq:16}
\end{equation} 
where ${\textbf{\emph{Y}}}_i={\textbf{\emph{D}}}_i\boldsymbol\mu_i$ and ${\textbf{\emph{X}}}_i={\textbf{\emph{D}}}_i\boldsymbol\alpha_i$.

\emph{Proof:} see Appendix.

\noindent$\textbf{{Theorem 2}}$

 \ \ The equivalence of the group-based sparse coding and the rank minimization problem is satisfied under the adaptive dictionary ${\textbf{\emph{D}}}_i$.

\emph{Proof:} see Appendix.

It can be similarly proven that the reweighted $\ell_1$-norm and $\ell_p$-norm  minimization are equivalent to the weighted nuclear norm minimization (WNNM) \cite{13} and the weighted Schatten $p$-norm minimization (WSNM) \cite{14}, respectively.

Note that the main difference between sparse coding and the rank minimization problem is that sparse coding has a dictionary learning operator and the rank minimization problem does not involve.
\subsection{Analyzing the sparsity of  group based on the nuclear norms minimization}
\begin{figure}[t]
\begin{minipage}[b]{1\linewidth}
  \centerline{\includegraphics[width=8cm]{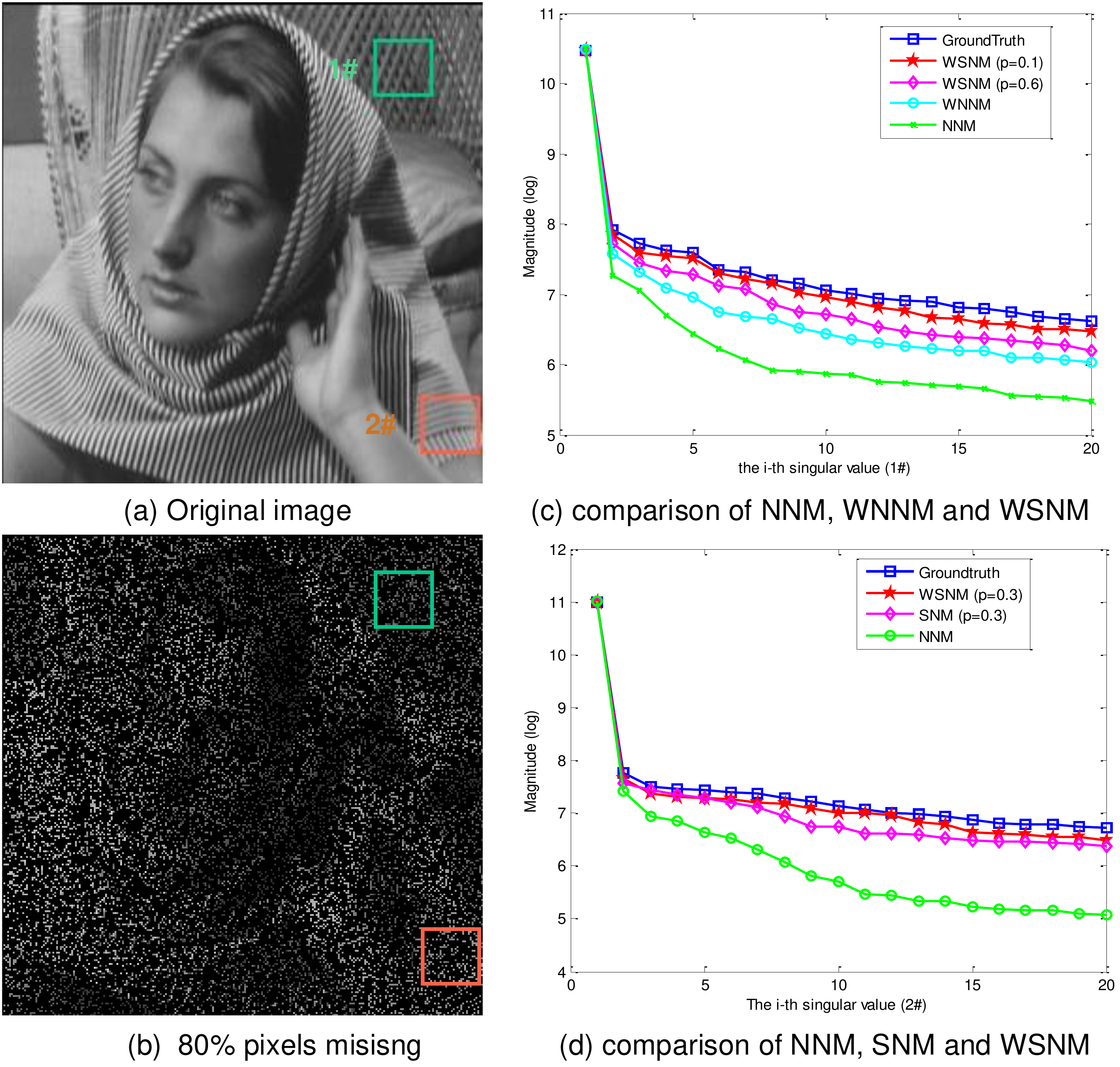}}
\end{minipage}
\caption{Analyzing the sparsity of each group based nuclear norms minimization.}
\label{fig:2}
\end{figure}

Based on \emph{Theorem 2}, the group-based sparse coding can be turned into the rank minimization problem. Now, four nuclear norms are used to constrain Eq.~\eqref{eq:2} to analyze the sparsity of each group, i.e., nuclear norm minimization (NNM) \cite{11}, the weighted nuclear norm minimization (WNNM) \cite{12}, Schatten $p$-norm minimization (SNM) \cite{13} and the weighted Schatten $p$-norm minimization (WSNM) \cite{14}. In these experiments, a gray image \emph{Barbara} is used as an example in the context of image inpainting, where 80\% pixels are damaged in Fig.~\ref{fig:2}(b). We generate two groups based on 1\# position and 2\# position which are shown in Fig.~\ref{fig:2}(a). As shown in Fig.~\ref{fig:2}(c) and Fig.~\ref{fig:2}(d), it can be seen that the singular values of WSNM result are the best approximation to the ground-truth in comparison with other methods. Therefore, based on $\emph{Theorem 2}$, WSNM can be equivalently transformed into solving the non-convex $\ell_p$-norm minimization to measure the sparsity  in group-based sparse coding.
\section{Image restoration using group-based sparse coding with non-convex $\ell_p$-norm minimization}
In this section, we verify the proposed scheme in the application of image restoration (IR). IR aims to reconstruct a high quality image $\textbf{\emph{X}}$ from its degraded observation $\textbf{\emph{Y}}$,
\begin{equation}
\textbf{\emph{Y}}=\textbf{\emph{H}}\textbf{\emph{X}} +\textbf{\emph{N}}
\label{eq:21}
\end{equation} 
where $\textbf{\emph{H}}$ is the degraded operator and $\textbf{\emph{N}}$ is usually assumed to be additive white Gaussian noise. In this paper, we will focus on image inpainting and image compressive sensing (CS) recovery.

In the scenario of IR, what we observed is the degraded image $\textbf{\emph{Y}}$ via Eq.~\eqref{eq:21}, and thus the goal is to employ  the proposed scheme to recover the original image $\textbf{\emph{X}}$ from $\textbf{\emph{Y}}$ by solving the following non-convex $\ell_p$-norm minimization problem,
  \begin{equation}
{{\boldsymbol\alpha}}=\arg\min\limits_{\boldsymbol\alpha}
\frac{1}{2}||{\textbf{\emph{Y}}}-\textbf{\emph{H}}{\textbf{\emph{D}}}{{\boldsymbol\alpha}}||_2^2+\lambda||{\textbf{\emph{w}}}{{\boldsymbol\alpha}}||_p
\label{eq:22}
\end{equation} 
\subsection{ADMM based for non-convex $\ell_p$-norm minimization}
 Solving the objective function of Eq.~\eqref{eq:22} is very difficult, since it is a large scale non-convex optimization problem. To make the proposed scheme tractable and robust, in this paper we present the alternating direction method of multipliers (ADMM) \cite{18} to solve Eq.~\eqref{eq:22}. Specifically, we introduce an auxiliary variable $\textbf{\emph{Z}}$ and Eq.~\eqref{eq:22} can be rewritten as
  \begin{equation}
{{\boldsymbol\alpha}}=\arg\min\limits_{\textbf{\emph{Z}}, \boldsymbol\alpha}
\frac{1}{2}||{\textbf{\emph{Y}}}-\textbf{\emph{H}}{\textbf{\emph{Z}}}||_2^2+\lambda||{\textbf{\emph{w}}}{{\boldsymbol\alpha}}||_p,\ \ s.t.\ \  \textbf{\emph{Z}}={\textbf{\emph{D}}}{{\boldsymbol\alpha}}
\label{eq:23}
\end{equation} 

By defining $f(\textbf{\emph{Z}})=\frac{1}{2}||{\textbf{\emph{Y}}}-\textbf{\emph{H}}{\textbf{\emph{Z}}}||_2^2$, and $g(\boldsymbol\alpha)=\lambda||{\textbf{\emph{w}}}{{\boldsymbol\alpha}}||_p$, we have
\begin{equation}
{\textbf{\emph{Z}}}^{t+1}=\arg\min\limits_{\textbf{\emph{Z}}}\frac{1}{2}||\textbf{\emph{Y}}-\textbf{\emph{H}}{\textbf{\emph{Z}}}||_2^2
+\frac{\rho}{2}||\textbf{\emph{Z}}-{{\textbf{\emph{D}}}{{\boldsymbol\alpha}}}^{t}-\textbf{\emph{C}}^{t}||_2^2
\label{eq:24}
\end{equation}
\begin{equation}
{\boldsymbol\alpha}^{t+1}=\arg\min\limits_{\boldsymbol\alpha}\lambda||{\textbf{\emph{w}}}{{\boldsymbol\alpha}}||_p
+\frac{\rho}{2}||\textbf{\emph{Z}}^{t+1}-{{\textbf{\emph{D}}}{{\boldsymbol\alpha}}}-\textbf{\emph{C}}^{t}||_2^2
\label{eq:25}
\end{equation}
and
\begin{equation}
{\textbf{\emph{C}}}^{t+1}={\textbf{\emph{C}}}^{t}-({\textbf{\emph{Z}}}^{t+1}-{{\textbf{\emph{D}}}{{\boldsymbol\alpha}}}^{t+1})
\label{eq:26}
\end{equation}\par
It can be seen that the minimization for Eq.~\eqref{eq:23} involves splitting two minimization sub-problems, i.e., $\textbf{\emph{Z}}$ and ${{{\boldsymbol\alpha}}}$ sub-problem. Next, we will show that there is an efficient solution to each sub-problem. To avoid confusion, the subscribe $t$ may be omitted for conciseness.
\subsubsection{${\textbf{\emph{Z}}}$ sub-problem}
Given ${{{\boldsymbol\alpha}}}$, the $\textbf{\emph{Z}}$ sub-problem denoted by Eq.~\eqref{eq:24} becomes
\begin{equation}
\min_{\textbf{\emph{Z}}}{\textbf{\emph{L}}}_1({{\textbf{\emph{Z}}}})=\min\limits_{\textbf{\emph{Z}}}\frac{1}{2}||\textbf{\emph{Y}}-\textbf{\emph{H}}{\textbf{\emph{Z}}}||_2^2
+\frac{\rho}{2}||\textbf{\emph{Z}}-{{\textbf{\emph{D}}}{{\boldsymbol\alpha}}}-\textbf{\emph{C}}||_2^2
\label{eq:27}
\end{equation}
Clearly, Eq.~\eqref{eq:27} has a closed-form solution and its solution can be expressed as
\begin{equation}
\hat{\textbf{\emph{Z}}}=({\textbf{\emph{H}}}^{T}{\textbf{\emph{H}}}+\rho{\textbf{\emph{I}}})^{-1}({\textbf{\emph{H}}}^{T}{\textbf{\emph{Y}}}+\rho({{\textbf{\emph{D}}}{{\boldsymbol\alpha}}}+\textbf{\emph{C}}))
\label{eq:28}
\end{equation} 
where ${\textbf{\emph{I}}}$ represents the identity matrix.
\subsubsection{${{{\boldsymbol\alpha}}}$ sub-problem}
Given ${\textbf{\emph{Z}}}$, similarity, according to Eq.~\eqref{eq:25}, the ${{{\boldsymbol\alpha}}}$ sub-problem can be written as
\begin{equation}
\min_{{{{\boldsymbol\alpha}}}}{\textbf{\emph{L}}}_2({{{{\boldsymbol\alpha}}}})=
\min\limits_{{{\boldsymbol\alpha}}}
\frac{1}{2}||{{\textbf{\emph{D}}}{{\boldsymbol\alpha}}}-\textbf{\emph{R}}||_2^2\\
+\frac{\lambda}{\rho}||{\textbf{\emph{w}}}{{\boldsymbol\alpha}}||_p
\label{eq:29}
\end{equation}
where  $\textbf{\emph{R}}=\textbf{\emph{Z}}-\textbf{\emph{C}}$.

However, due to the complex structure of $||{\textbf{\emph{w}}}{{\boldsymbol\alpha}}||_p$, it is difficult to solve Eq.~\eqref{eq:29}, Let $\textbf{\emph{X}}={{\textbf{\emph{D}}}{{\boldsymbol\alpha}}}$, Eq.~\eqref{eq:29} can be rewritten as
\begin{equation}
\min_{{{{\boldsymbol\alpha}}}}{\textbf{\emph{L}}}_2({{{{\boldsymbol\alpha}}}})=
\min\limits_{{{\boldsymbol\alpha}}}
\frac{1}{2}||{{\textbf{\emph{X}}}}-\textbf{\emph{R}}||_2^2\\
+\frac{\lambda}{\rho}||{\textbf{\emph{w}}}{{\boldsymbol\alpha}}||_p
\label{eq:30}
\end{equation}

To enable a tractable solution of Eq.~\eqref{eq:30}, in this paper, a general assumption is made, with which even a closed form can be achieved. Specifically, $\textbf{\emph{R}}$ can be regarded as some type of noisy observation of $\textbf{\emph{X}}$, and then the assumption is made that each element of
$\textbf{\emph{E}}=\textbf{\emph{X}}-\textbf{\emph{R}}$ follows an independent zero-mean distribution with variance ${
\sigma}^{2}$.  The following conclusion can be proved by this assumption.

\noindent$\textbf{Theorem 3}$ \ \ Define $\textbf{\emph{X}},\textbf{\emph{R}}\in\Re^{N}$, ${\textbf{\emph{X}}}_i$, ${\textbf{\emph{R}}}_i$, and ${\textbf{\emph{e}}}{(j)}$ as each element of  error vector ${\textbf{\emph{e}}}$, where $\textbf{\emph{e}}=\textbf{\emph{X}}-\textbf{\emph{R}},  j=1,...,N$. Assume that ${\textbf{\emph{e}}}{(j)}$ follows an independent zero mean distribution with variance ${
\sigma}^{2}$, and thus for any $\varepsilon>0$, we can represent the relationship between $\frac{1}{N}||\textbf{\emph{X}}-\textbf{\emph{R}}||_2^2$ and ${\frac{1}{K}}\sum_{i=1}^n||{\textbf{\emph{X}}}_i-{\textbf{\emph{R}}}_i||_2^2$  by the following property,
\begin{equation}
\lim_{{N\rightarrow\infty}\atop{K\rightarrow\infty}}{\textbf{\emph{P}}}{\{|\frac{1}{N}||\textbf{\emph{X}}-\textbf{\emph{R}}||_2^2
-{\frac{1}{K}}\sum\nolimits_{i=1}^n||{\textbf{\emph{X}}}_i-{\textbf{\emph{R}}}_i||_F^2|<\varepsilon\}}=1
\label{eq:31}
\end{equation}
where ${\textbf{\emph{P}}}(\bullet)$  represents the probability and ${\emph{K}}=\emph{d}\times\emph{k}\times\emph{n}$. The detailed proof of $\emph{Theorem 3}$ is given in  supplemental matarial.

Based on $\emph{Theorem 3}$, we have the following equation with a very large probability (restricted 1) at each iteration,
\begin{equation}
\frac{1}{N}||\textbf{\emph{X}}-\textbf{\emph{R}}||_2^2
={\frac{1}{K}}\sum\nolimits_{i=1}^n||{\textbf{\emph{X}}}_i-{\textbf{\emph{R}}}_i||_F^2
\label{eq:32}
\end{equation}

Based on Eqs.~\eqref{eq:30} and~\eqref{eq:32}, we have
\begin{equation}
\begin{aligned}
&\min\limits_{{{\boldsymbol\alpha}}}\frac{1}{2}{||\textbf{\emph{X}}-\textbf{\emph{R}}||_2^2}
 +\frac{\lambda}{\rho}||{\textbf{\emph{w}}}_{{\boldsymbol\alpha}}||_p\\
&=\min\limits_{{{{\boldsymbol\alpha}}}_i}(\sum\nolimits_{i=1}^n\frac{1}{2}||{\textbf{\emph{X}}}_i-{\textbf{\emph{R}}}_i||_F^2
 +{{\tau}}_i||{{{\emph{w}}}}_i{{{\boldsymbol\alpha}}}_i||_p)\\
 &=\min\limits_{{{{\boldsymbol\alpha}}}_i}(\sum\nolimits_{i=1}^n\frac{1}{2}||{\textbf{\emph{R}}}_i-{{\textbf{\emph{D}}}_i{{{\boldsymbol\alpha}}}_i}||_F^2
 +{{\tau}}_i||{{{\emph{w}}}}_i{{{\boldsymbol\alpha}}}_i||_p)\\
\end{aligned}
\label{eq:33}
\end{equation}
where ${{\tau}}_i={{{\lambda}}_i{\emph{K}}}/{\rho{\emph{N}}}$ and ${{\textbf{\emph{D}}}_i}$ is a dictionary. Clearly, Eq.~\eqref{eq:33} can be viewed as a sparse coding problem by solving $n$ sub-problems for all the group ${\textbf{\emph{X}}}_i$.  Based on  \emph{Theorem 1}, Eq.~\eqref{eq:33} can be rewritten as:
\begin{equation}
{\hat{\boldsymbol\alpha}}_i=\min\limits_{{{{\boldsymbol\alpha}}}_i}\sum\nolimits_{i=1}^n\frac{1}{2}||{{{{\boldsymbol\gamma}}}_i}-{{{{\boldsymbol\alpha}}}_i}||_2^2
 +{{\tau}}_i{{{\emph{w}}}}_i||{{{\boldsymbol\alpha}}}_i||_p
\label{eq:34}
\end{equation}
where ${\textbf{\emph{R}}}_i={{\textbf{\emph{D}}}_i{{{\boldsymbol\gamma}}}_i}$ and
${\textbf{\emph{X}}}_i={{\textbf{\emph{D}}}_i{{{\boldsymbol\alpha}}}_i}$.

 To obtain the solution of Eq.~\eqref{eq:34} effectively, in this paper, the generalized soft-thresholding (GST) algorithm \cite{19} is used to solve Eq.~\eqref{eq:34}. Therefore, a closed-form solution of  Eq.~\eqref{eq:34} can be computed as
 \begin{equation}
{\hat{\boldsymbol\alpha}}_i={{\emph{GST}}}({{{\boldsymbol\gamma}}}_i, {{\tau}}_i{{{\emph{w}}}}_i, p)
\label{eq:35}
\end{equation}

For more details about the GST algorithm, please refer to \cite{19}. For each weight ${{{\emph{w}}}}_i$, large values of sparse coefficient ${{{\boldsymbol\alpha}}}_i$ usually transmit major edge and texture information \cite{5}. This implies that to reconstruct ${\textbf{\emph{X}}}_i$ from its degraded one, we should shrink large values less, while shrinking smaller ones more, and thus we have ${{{\emph{w}}}}_i= 1/(|{{{\boldsymbol\gamma}}}_i|+\epsilon)$, where $\epsilon$ is a small constant. Inspired by \cite{20}, the regularization parameter $\lambda_i$ of each group ${\textbf{\emph{R}}}_i$ is set as: $\lambda_i=2\sqrt{2}\sigma^2/(\delta_i+\varepsilon)$, where $\delta_i$ denotes the estimated variance of ${{{\boldsymbol\gamma}}}_i$, and $\varepsilon$ is a small constant.
 After solving the two sub-problems, we summarize the overall algorithm for Eq.~\eqref{eq:23} in \emph{Table 1}.
  \begin{table}[t]
\caption{ADMM method for Eq.~\eqref{eq:23}}
\centering  
\begin{tabular}{lccc}  
\hline  
$\textbf{Input:}$ the observed image $\textbf{\emph{Y}}$ and the measurement matrix $\textbf{\emph{H}}.$\\
  $\textbf{Initialization:}\ t$, ${{\textbf{\emph{C}}}}$,
  ${{\textbf{\emph{Z}}}}$, ${{\boldsymbol\alpha}}$,
  $\emph{I}$, $\emph{d}$, $\emph{k}$, $\rho$, $p$, $\sigma$, $\epsilon$, $\varepsilon$; \\
  \textbf{Repeat} \\
    \qquad \qquad Update ${{\textbf{\emph{Z}}}}^{t+1}$ by Eq.~\eqref{eq:28}; \\
  \qquad \qquad ${\textbf{\emph{R}}}^{{t}+1}={\textbf{\emph{Z}}}^{{t}+1}-{\textbf{\emph{C}}}^{t}$;\\
  $\textbf{For}$  \qquad Each group ${\textbf{\emph{R}}}_{i}$;\\
  \qquad \qquad Construct dictionary ${\textbf{\emph{D}}}_i$ by computing Eq.~\eqref{eq:9};\\
   \qquad \qquad Update ${\lambda_i}^{t+1}$ by computing  $\lambda_i=2\sqrt{2}\sigma^2/\delta_i+\varepsilon$;\\
   \qquad \qquad Update $\tau_i^{t+1}$  computing by ${{\tau}}_i={{{\lambda}}_i{\emph{K}}}/{\rho{\emph{N}}}$;\\
  \qquad \qquad Update ${{\emph{w}}}_i^{t+1}$  computing by ${{{\emph{w}}}}_i= {{\tau}}_i/|{{{\boldsymbol\gamma}}}_i|+\epsilon$;\\
    \qquad \qquad Update ${\boldsymbol\alpha}_i^{t+1}$  computing by Eq.~\eqref{eq:35};\\
  $\textbf{End For}$ \\
  \qquad \qquad Update ${\textbf{\emph{D}}}^{t+1}$ by concatenating all ${\textbf{\emph{D}}}_i$ ;\\
    \qquad \qquad Update ${\boldsymbol\alpha}^{t+1}$ by concatenating all ${\boldsymbol\alpha}_i$ ;\\
      \qquad \qquad Update ${{\textbf{\emph{C}}}^{t+1}}$ by computing Eq.~\eqref{eq:26} ;\\

    \qquad \qquad  $t\leftarrow t+1$;\\
     $\textbf{Until}$\\ \qquad maximum iteration number is reached.\\
     $\textbf{Output:}$\\ \qquad The final restored image $\hat{\textbf{\emph{X}}}={\textbf{\emph{D}}}{\hat{\boldsymbol\alpha}}.$\\
\hline
\end{tabular}
\end{table}
\begin{figure}[t]
\begin{minipage}[b]{1\linewidth}
  \centerline{\includegraphics[width=7.5cm]{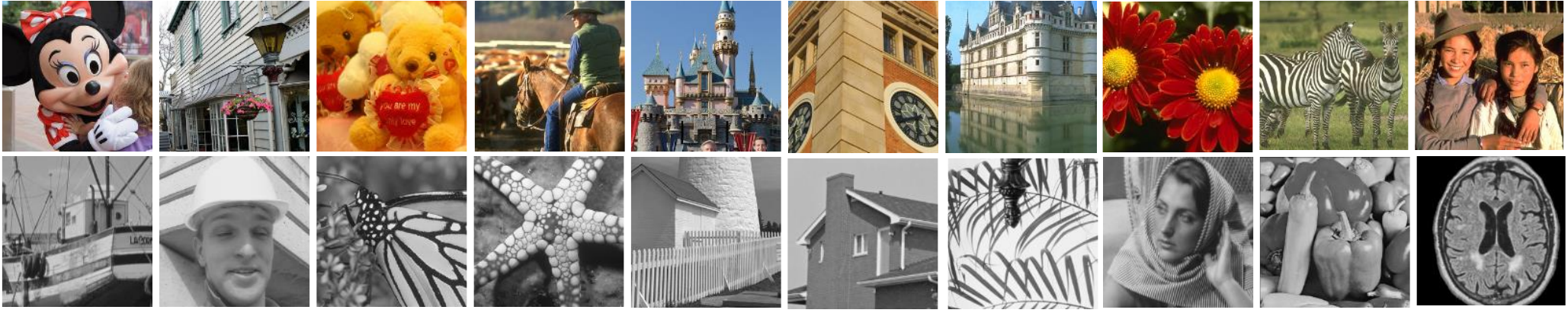}}
\end{minipage}
\caption{All test images.}
\label{fig:5}
\end{figure}
\begin{figure}[t]
\begin{minipage}[b]{1\linewidth}
  \centerline{\includegraphics[width=9cm]{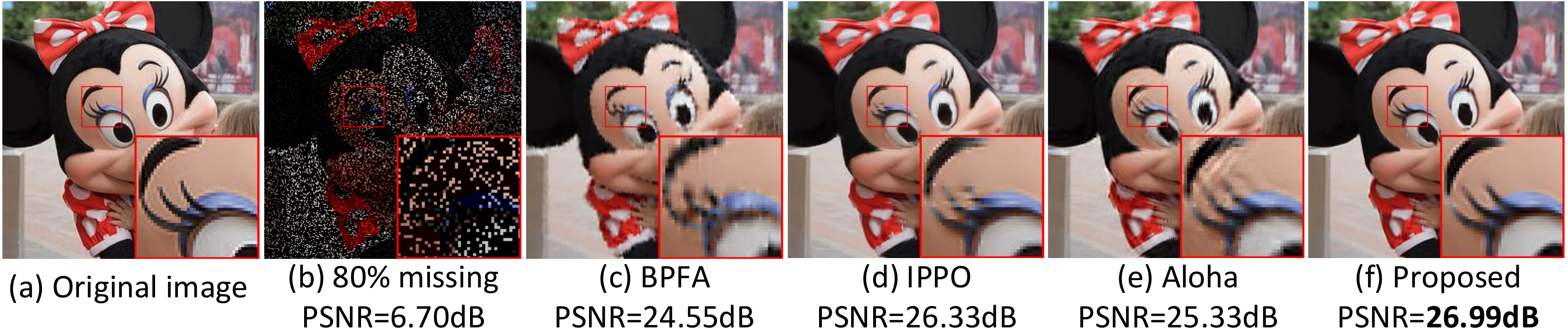}}
\end{minipage}
\caption{Visual comparison of \emph{Mickey} by image inpainting with 80\% missing pixels.}
\label{fig:3}
\end{figure}
\begin{figure}[!htbp]
\begin{minipage}[b]{1\linewidth}
  \centerline{\includegraphics[width=9cm]{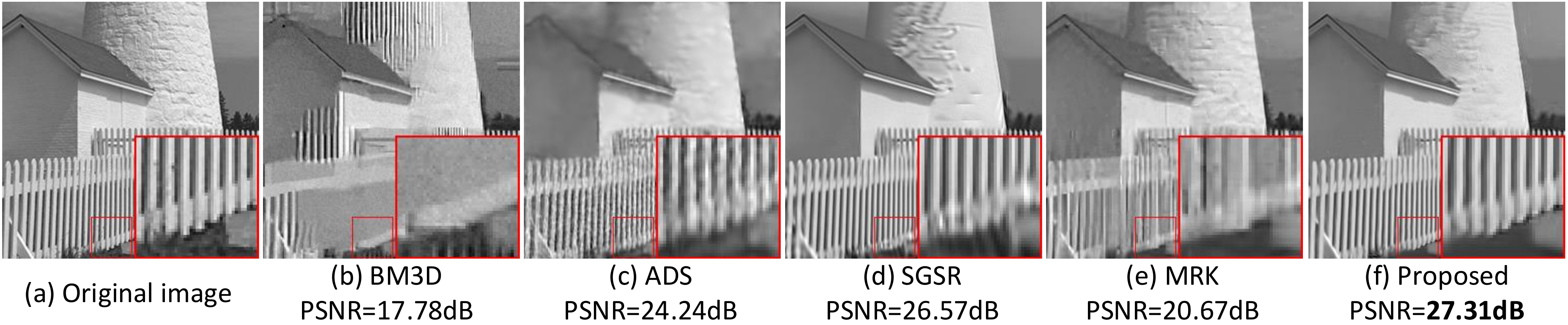}}
\end{minipage}
\caption{Visual comparison  of \emph{Fence} by CS recovery with 0.1$N$ measurements.}
\label{fig:4}
\end{figure}
\begin{table}[t]
\caption{Average PSNR results (dB) of image inpainting}
\centering  
\begin{tabular}{|c|c|c|c|c|c|c|}
\hline
\multirow{1}{*}{\textbf{{Method}}}&{\textbf{{80\%}}}&{\textbf{{70\%}}}&{\textbf{{60\%}}}&{\textbf{{50\%}}}
&{\textbf{{Inlay text}}}\\
 \hline
  \multirow{1}{*}{BPFA} & 24.35 & 26.26 & 28.03 & 29.83 & 31.42  \\
\hline
  \multirow{1}{*}{IPPO} & 25.40 & 27.34 & 29.19 & 30.93 & 32.90 \\
\hline
 \multirow{1}{*}{Aloha} & 25.13 & 26.92 & 28.51 & 30.13 & 31.12\\
 \hline
 \multirow{1}{*}{Proposed} & \textbf{25.91} & \textbf{28.04} & \textbf{29.98} & \textbf{31.95} & \textbf{33.58}\\
  \hline
\end{tabular}
\end{table}
\begin{table}[t]
\caption{Average PSNR results (dB) of image CS recovery}
\centering  
\begin{tabular}{|c|c|c|c|c|c|c|}
\hline
\multirow{1}{*}{\textbf{{Ratio}}}&{\textbf{{BM3D}}}&{\textbf{{ADS}}}&{\textbf{{SGSR}}}&{\textbf{{MRK}}}
&{\textbf{{Proposed}}}\\
 \hline
  \multirow{1}{*}{0.1} & 24.23 & 27.65 & 27.93 & 27.14 & \textbf{28.89}  \\
\hline
  \multirow{1}{*}{0.2} & 30.60 & 32.42 & 32.36 & 31.22 & \textbf{33.50} \\
\hline
 \multirow{1}{*}{0.3} & 34.11 & 35.68 & 34.87 & 34.20 & \textbf{36.50}\\
  \hline
\end{tabular}
\end{table}
\section{Experimental Results}
 In this section, we report our experimental results in the applications of image inpainting and image CS recovery.  All the experimental images are shown in Fig.~\ref{fig:5}. 

 In image inpainting, two interesting examples of image inpainting with different masks are conducted, i.e., image restoration from partial random samples and text inlayed sample. The parameters are set as follows. The size of each patch $\sqrt{d}\times \sqrt{d}$ is set to be $8\times 8$ and $10\times 10$ for partial random samples and text inlayed, respectively. Similar patch numbers $k =60$, $\emph{I}=25$, $\sigma=\sqrt{2}$, $\epsilon=0.1$, $\varepsilon=0.3$. ($\rho$, $p$) are set to (0.0003, 0.45), (0.0003, 0.45), (0.03, 1), (0.04, 1) and (0.06, 0.95) when 80\%, 70\%, 60\%, 50\% pixels missing and text inlayed, respectively.

 We compared the proposed scheme  with three other competing methods: BPFA method \cite{2}, IPPO method \cite{21}, Aloha method \cite{22}. Table 2 lists the average PSNR comparison results for a  collection  of 10 color images among three competing methods. The visual comparison of the image inpainting methods is shown in Fig.~\ref{fig:3}. It can be seen that BPFA could not reconstruct sharp edges and  fine details. The IPPO and Aloha methods produced images with a much better visual quality than BPFA method, but still suffered from some undesirable artifacts, such as the ringing effects. The proposed scheme  not only preserved sharper edges and finer details, but eliminated the ringing effects.

 In image CS recovery,  we generated the CS measurements at the block level by utilizing a Gaussian random projection matrix to test images, i.e., the CS recovery with block size $32\times 32$. The parameters are set as follows. The size of each patch $\sqrt{d}\times \sqrt{d}$ is set to be $7\times 7$. Similar patch numbers $k =60$, $\emph{I}=20$, $\sigma=\sqrt{2}$, $\epsilon=0.1$, $\varepsilon=0.4$. ($\rho$, $p$) are set to (0.0001, 0.65), (0.0005, 0.5) and (0.05, 1) when 0.1$N$, 0.2$N$ and 0.3$N$ measurements, respectively.

 We have compared the proposed method based image CS recovery against four other competing approaches including BM3D method \cite{23}, ADS method \cite{4}, SGSR method \cite{25} and MRK method \cite{26}. The average PSNR results are shown in Table 3. The average gain of the proposed scheme over BM3D, ADS, SGSR and MRK methods can be as much as 3.31dB, 1.04dB, 1.24dB and 2.11dB, respectively. The visual comparisons of the image CS recovery are shown in Fig.~\ref{fig:4}. It can be seen that the BM3D, ADS, SGSR, and MRK methods still suffer from some undesirable artifacts or over-smooth phenomena. By contrast, the proposed method not only removes most of the visual artifacts, but also preserves large-scale sharp edges and small-scale fine image details more effectively.

\section{Conclusion}
 This paper proposed to analyze the group sparsity based on the rank minimization methods. An adaptive dictionary learning method for each group is designed. We prove the equivalence of the group-based sparse coding and the rank minimization problem, and thus the sparse coefficient of each group are measured by computing the singular values of each group. Four nuclear norms are used to analyze the sparsity of  each group and the solution of the weighted Schatten $p$-norm minimization (WSNM) is the best approximation to real singular values of each group.   Therefore,  WSNM can be equivalently transformed into a non-convex $\ell_p$-norm minimization problem in group-based sparse coding. To make the proposed scheme tractable and robust, we utilize the alternating direction method of multipliers (ADMM)  to solve the $\ell_p$-norm minimization problem. Experimental results  on image inpainting and image CS recovery have demonstrated that the proposed scheme achieves significant performance improvements over the current state-of-the-art methods.

 \section{Appendix}
 Consider the SVD of a matrix $\textbf{\emph{P}}\in\Re^{n_1\times n_2}$ of rank $r$
\begin{equation}
\textbf{\emph{P}}= \textbf{\emph{U}}\boldsymbol\Sigma\textbf{\emph{V}}^T, \boldsymbol\Sigma ={\rm diag}(\{\sigma_i\}_{1\leq i\leq r})
\label{eq:36}
\end{equation} 
where $\textbf{\emph{U}}\in\Re^{n_1 \times r}$ and $\textbf{\emph{V}}\in\Re^{n_2 \times r}$ are orthogonal matrices, respectively. $\sigma_i$ is the $i$-th singular value of $\textbf{\emph{P}}$. For each $\tau\geq0$, the soft-thresholding operator $\mathcal{D}_\tau$ is defined as
\begin{equation}
\mathcal{D}_\tau(\textbf{\emph{P}})= \textbf{\emph{U}} \mathcal{D}_\tau(\boldsymbol\Sigma)\textbf{\emph{V}}^T,\ \ \ \mathcal{D}_\tau(\boldsymbol\Sigma)= {\rm soft}(\sigma_i,\tau)
\label{eq:37}
\end{equation} 
\noindent$\textbf{{Lemma 2}}$

\ \ \ \ \ \ \ \ \ \ \  For each $\tau\geq0$, and $\textbf{\emph{Q}}\in\Re^{n_1\times n_2}$ ,the singular value shrinkage operator Eq.~\eqref{eq:37} satisfies
\begin{equation}
\mathcal{D}_\tau(\textbf{\emph{Q}})= \arg\min_{\textbf{\emph{P}}} \{\frac{1}{2}||{\textbf{\emph{P}}}-{\textbf{\emph{Q}}}||_F^2+\tau||{\textbf{\emph{P}}}||_*\}
\label{eq:38}
\end{equation} 

\noindent\emph{Proof:}\ \  Due to the strict convex of the function $h(\textbf{\emph{P}}): = \frac{1}{2}||{\textbf{\emph{P}}}-{\textbf{\emph{Q}}}||_F^2+\tau||{\textbf{\emph{P}}}||_*$, it is easy to observe that there exists an unique minimizer, and thus we are only required to prove that it is equal to $\mathcal{D}_\tau({\textbf{\emph{Q}}})$. To this end, recall the definition of a subgradient of a convex function $f: \Re^{n_1 \times n_2}\rightarrow \Re$. We say that $\textbf{\emph{Z}}$ is a subgradient of $f$ at ${\textbf{\emph{P}}}_0$, denoted  $\textbf{\emph{Z}}\in\partial f({\textbf{\emph{P}}}_0)$, if
\begin{equation}
f({\textbf{\emph{P}}}) \geq f({\textbf{\emph{P}}}_0)+ \langle\textbf{\emph{Z}},{\textbf{\emph{P}}}-{\textbf{\emph{P}}}_0\rangle
\label{eq:39}
\end{equation} 
for all $\textbf{\emph{P}}$. Now $\hat{\textbf{\emph{P}}}$ minimizes $h$ if and only if it satisfies the following optimal condition, i.e.,
\begin{equation}
\textbf{0} \in \hat{\textbf{\emph{P}}}-{\textbf{\emph{Q}}} + \tau\partial||\hat{\textbf{\emph{P}}}||_*
\label{eq:40}
\end{equation} 
where $\partial||\hat{\textbf{\emph{P}}}||_*$ is the set of subgradients of the nuclear norm. Let matrix ${\textbf{\emph{P}}} \in \Re^{n_1 \times n_2}$ and its SVD be $\textbf{\emph{U}}\boldsymbol\Sigma \textbf{\emph{V}}^T$. It is known from \cite{27} that the subgradient of $||{\textbf{\emph{P}}}||_*$ can be derived as
\begin{equation}
\partial||{\textbf{\emph{P}}}||_*=\{\textbf{\emph{U}}\textbf{\emph{V}}^T+ \textbf{\emph{S}}:\ \ \textbf{\emph{S}}\in \Re^{n_1 \times n_2}, \textbf{\emph{U}}^T \textbf{\emph{S}}=0, \textbf{\emph{S}}\textbf{\emph{V}}=0, ||\textbf{\emph{S}}||_2\leq1   \}
\label{eq:41}
\end{equation} 

We set $\hat{\textbf{\emph{P}}} = \mathcal{D}_\tau({\textbf{\emph{Q}}})$, to show that $\hat{\textbf{\emph{P}}}$ satisfies Eq.~\eqref{eq:40}, we rewritten the SVD of $\textbf{\emph{Q}}$ as $\textbf{\emph{Q}}=\textbf{\emph{U}}_0\boldsymbol\Sigma_0\textbf{\emph{V}}_0^T+ \textbf{\emph{U}}_1\boldsymbol\Sigma_1\textbf{\emph{V}}_1^T$, where $\textbf{\emph{U}}_0, \textbf{\emph{V}}_0  ({\rm resp}.  \textbf{\emph{U}}_1, \textbf{\emph{V}}_1)$ are the singular vectors associated with singular values greater than $\tau$ (resp. smaller than or equal to $\tau$). With these notations, we have
\begin{equation}
\hat{\textbf{\emph{P}}}= \textbf{\emph{U}}_0 (\boldsymbol\Sigma_0-\tau \textbf{\emph{I}})\textbf{\emph{V}}_0^T
\label{eq:42}
\end{equation} 
Therefore,
\begin{equation}
\textbf{\emph{Q}}-\hat{\textbf{\emph{P}}}= \tau (\textbf{\emph{U}}_0\textbf{\emph{V}}_0^T+\textbf{\emph{S}}), \ \ \textbf{\emph{S}}=\tau^{-1}\textbf{\emph{U}}_1\boldsymbol\Sigma_1 \textbf{\emph{V}}_1^T
\label{eq:43}
\end{equation} 

By definition, $\textbf{\emph{U}}_0^T \textbf{\emph{S}}=0, \textbf{\emph{S}}\textbf{\emph{V}}_0=0.$ Since the diagonal elements of $\boldsymbol\Sigma_1$ have magnitudes bounded by $\tau$, we also have $||\textbf{\emph{S}}||_2\leq1.$ Therefore, $\textbf{\emph{Q}}-\hat{\textbf{\emph{P}}}\in\tau\partial||\hat{\textbf{\emph{P}}}||_*$, which concludes the proof.

\noindent$\textbf{{Theorem 1}}$
\begin{equation}
||{\textbf{\emph{Y}}}_i-{\textbf{\emph{X}}}_i||_F^2=||\boldsymbol\mu_i-\boldsymbol\alpha_i||_2^2
\label{eq:44}
\end{equation} 
where ${\textbf{\emph{Y}}}_i={\textbf{\emph{D}}}_i\boldsymbol\mu_i$ and ${\textbf{\emph{X}}}_i={\textbf{\emph{D}}}_i\boldsymbol\alpha_i$.

\noindent\emph{Proof:} \ \ Since the adaptive dictionary ${\textbf{\emph{D}}}_i$ is constructed by Eq.~\eqref{eq:9}, and  the unitary property of ${\textbf{\emph{U}}}_i$ and ${\textbf{\emph{V}}}_i$, we have
\begin{equation}
\begin{aligned}
&||{\textbf{\emph{Y}}}_i-{\textbf{\emph{X}}}_i||_F^2=||{\textbf{\emph{D}}}_i(\boldsymbol\mu_i-\boldsymbol\alpha_i)||_F^2
=||{\textbf{\emph{U}}}_i{\rm diag}(\boldsymbol\mu_i-\boldsymbol\alpha_i){\textbf{\emph{V}}}_i||_F^2\\
&= {\rm trace}({\textbf{\emph{U}}}_i{\rm diag}({{{{\boldsymbol\mu}}_{\textbf{\emph{G}}}}_i}-{{{{\boldsymbol\alpha}}_{\textbf{\emph{G}}}}_i}){\textbf{\emph{V}}}_i
{\textbf{\emph{V}}}_{i}^T{\rm diag}(\boldsymbol\mu_i-\boldsymbol\alpha_i){\textbf{\emph{U}}}_{i}^T)\\
&= {\rm trace}({\textbf{\emph{U}}}_i{\rm diag}(\boldsymbol\mu_i-\boldsymbol\alpha_i)
{\rm diag}(\boldsymbol\mu_i-\boldsymbol\alpha_i){\textbf{\emph{U}}}_{i}^T)\\
&={\rm trace}({\rm diag}(\boldsymbol\mu_i-\boldsymbol\alpha_i)
{\textbf{\emph{U}}}_i{\textbf{\emph{U}}}_{i}^T{\rm diag}(\boldsymbol\mu_i-\boldsymbol\alpha_i))\\
&={\rm trace}({\rm diag}(\boldsymbol\mu_i-\boldsymbol\alpha_i)
{\rm diag}(\boldsymbol\mu_i-\boldsymbol\alpha_i))\\
&=||\boldsymbol\mu_i-\boldsymbol\alpha_i||_2^2
\end{aligned}
\label{eq:45}
\end{equation} 

This completes the proof of the theorem.

\noindent$\textbf{{Theorem 2}}$

\ \ \ \ \ \ \ \ \ \ \  The equivalence of the group-based sparse coding and the rank minimization problem is satisfied under the dictionary ${\textbf{\emph{D}}}_i$.

\noindent\emph{Proof:} On the basis of $\emph{{Theorem 1}}$, we have
  \begin{equation}
  \begin{aligned}
{{\boldsymbol\alpha}}_i&=\arg\min\limits_{\boldsymbol\alpha_i}
\{\frac{1}{2}||{\textbf{\emph{Y}}}_i-{\textbf{\emph{D}}}_i{\boldsymbol\alpha}_i||_F^2+\lambda||{\boldsymbol\alpha}_i||_1\}\\
&=\arg\min\limits_{\boldsymbol\alpha_i}\{\frac{1}{2}||{\boldsymbol\mu}_i-{\boldsymbol\alpha}_i||_2^2+\lambda||{\boldsymbol\alpha}_i||_1\}
\end{aligned}
\label{eq:46}
\end{equation} 

Thus, based on $\emph{{Lemma 1}}$, we have
\begin{equation}
\begin{aligned}
{{\boldsymbol\alpha}}_i & ={\rm soft}({{\boldsymbol\mu}}_i,\lambda)= {\rm sgn}({{\boldsymbol\mu}}_i,\lambda). {\rm max}(abs({{\boldsymbol\mu}}_i)-\lambda,0)\\
\end{aligned}
\label{eq:47}
\end{equation}

Obviously, according to Eqs.~\eqref{eq:8} and ~\eqref{eq:9}, we have
\begin{equation}
\begin{aligned}
&{\textbf{\emph{D}}}_i{{\boldsymbol\alpha}}_i= \sum\nolimits_{j=1}^m {{\rm soft}({\boldsymbol\mu}}_{i,j},\lambda){\textbf{\emph{d}}}_{i,j}\\
&=\sum\nolimits_{j=1}^m {{\rm soft}({\boldsymbol\mu}}_{i,j},\lambda){\textbf{\emph{u}}}_{i,j}{\textbf{\emph{v}}}_{i,j}^T \\
&={\textbf{\emph{U}}}_i\mathcal{D}_\tau(\boldsymbol\Sigma_i){\textbf{\emph{V}}}_i^T
\end{aligned}
\label{eq:48}
\end{equation} 
where ${{\boldsymbol\alpha}}_{i,j}$ represents the $j$-th element of the $i$-th group sparse coefficient ${{\boldsymbol\alpha}}_{i}$, and $\boldsymbol\Sigma_i$ is the  singular value matrix of the $i$-th group ${\textbf{\emph{Y}}}_i$.

Thus, based on \emph{{Lemma 2}}, we prove that the group-based sparse coding (Eq.~\eqref{eq:46}) is equivalent to the rank minimization problem (Eq.~\eqref{eq:38}).

Note that the main difference between sparse coding and the rank minimization problem is that sparse coding has a dictionary learning operator and the rank minimization problem does not involve.

\noindent$\textbf{Theorem 3}$

\ \ \ \ \ \ \ \ \ \ \ Define $\textbf{\emph{X}},\textbf{\emph{R}}\in\Re^{N}$, ${\textbf{\emph{X}}}_i$, ${\textbf{\emph{R}}}_i$, and ${\textbf{\emph{e}}}{(j)}$ as each element of  error vector ${\textbf{\emph{e}}}$, where $\textbf{\emph{e}}=\textbf{\emph{X}}-\textbf{\emph{R}},  j=1,...,N$. Assume that ${\textbf{\emph{e}}}{(j)}$ follows an independent zero mean distribution with variance ${
\sigma}^{2}$, and thus for any $\varepsilon>0$, we can represent the relationship between $\frac{1}{N}||\textbf{\emph{X}}-\textbf{\emph{R}}||_2^2$ and ${\frac{1}{K}}\sum_{i=1}^n||{\textbf{\emph{X}}}_i-{\textbf{\emph{R}}}_i||_2^2$  by the following property, namely,
\begin{equation}
\lim_{{N\rightarrow\infty}\atop{K\rightarrow\infty}}{\textbf{\emph{P}}}{\{|\frac{1}{N}||\textbf{\emph{X}}-\textbf{\emph{R}}||_2^2
-{\frac{1}{K}}\sum\nolimits_{i=1}^n||{\textbf{\emph{X}}}_i-{\textbf{\emph{R}}}_i||_F^2|<\varepsilon\}}=1
\label{eq:49}
\end{equation}
where ${\textbf{\emph{P}}}(\bullet)$  represents the probability and ${\emph{K}}=\emph{d}\times\emph{k}\times\emph{n}$.

\noindent\emph{Proof:}\ \ Owing to the assumption that ${\textbf{\emph{e}}}{(\textbf{\emph{j}})}$  follows an independent zero mean distribution with variance  $\sigma^2$, namely, $\textbf{\emph{E}}[{\textbf{\emph{e}}}{(\textbf{\emph{j}})}]=0$ and $\textbf{\emph{Var}}[{\textbf{\emph{e}}}{(\textbf{\emph{j}})}]=\sigma^2$. Thus, it can be deduced that each ${\textbf{\emph{e}}}{(\textbf{\emph{j}})}^2$  is also independent, and the meaning of each  ${\textbf{\emph{e}}}{(\textbf{\emph{j}})}^2$  is:
\begin{equation}
\textbf{\emph{E}}[{\textbf{\emph{e}}}{(\textbf{\emph{j}})}^2]=\textbf{\emph{Var}}[{\textbf{\emph{e}}}{(\textbf{\emph{j}})}]
+[\textbf{\emph{E}}[{\textbf{\emph{e}}}{(\textbf{\emph{j}})}]]^2=\sigma^2, \emph{j}=1,2,...,\emph{N}
\label{eq:50}
\end{equation}\par

By invoking the $\emph{law of Large numbers}$ in probability theory, for any $\epsilon>0$, it leads to
$\lim\limits_{\emph{N}\rightarrow\infty}{\textbf{\emph{P}}}\{|\frac{1}{\emph{N}}\Sigma_{\emph{j}
=1}^{\emph{N}}{\textbf{\emph{e}}}{(\textbf{\emph{j}})}^2-\sigma^2|<\frac{\epsilon}{2}\}=1$, namely,
\begin{equation}
\lim\limits_{\emph{N}\rightarrow\infty}{\textbf{\emph{P}}}\{|\frac{1}{\emph{N}}
||\textbf{\emph{X}}-\textbf{\emph{R}}||_2^2-\sigma^2|<\frac{\epsilon}{2}\}=1
\label{eq:51}
\end{equation}\par
Next, we denote the concatenation of all the groups ${\textbf{\emph{X}}}_i$  and ${\textbf{\emph{R}}}_i,\ \emph{i}=1,2,...,\emph{n}$, by ${\textbf{\emph{X}}}$ and ${\textbf{\emph{R}}}$, respectively. Meanwhile, we denote the error of each element of  ${\textbf{\emph{X}}}-{\textbf{\emph{R}}}$ by ${\textbf{\emph{e}}}{\emph{(k)}},\ \emph{k}=1,2,...,\emph{K}$. We have also denote ${\textbf{\emph{e}}}{\emph{(k)}}$  following an independent zero mean distribution with variance $\sigma^2$.\par
Therefore, the same process applied to ${\textbf{\emph{e}}}{\emph{(k)}}^2$ yields $\lim\limits_{\emph{N}\rightarrow\infty}{\textbf{\emph{P}}}\{|\frac{1}{\emph{N}}\Sigma_{\emph{k}
=1}^{\emph{N}}{\textbf{\emph{e}}}{({\emph{k}})}^2-\sigma^2|<\frac{\epsilon}{2}\}=1$, i.e.,

\begin{equation}
\lim\limits_{\emph{N}\rightarrow\infty}{\textbf{\emph{P}}}\{|\frac{1}{\emph{N}}\Sigma_{\emph{i}
=1}^{\emph{n}}||{\textbf{\emph{X}}}_{i}-{\textbf{\emph{R}}}_{i}||_2^2-\sigma^2|<\frac{\epsilon}{2}\}=1
\label{eq:52}
\end{equation}\par
Obviously, considering Eqs.~\eqref{eq:51} and \eqref{eq:52} together, we can prove Eq.~\eqref{eq:49}.
 \section{Acknowledge}
The authors would like to thank Dr. Jian Zhang \footnote{ http://124.207.250.90/staff/zhangjian/} of Peking University  for his help. In addition, this work was supported by the NSFC (61571220, 61462052, 61502226) and the open research fund of National Mobile Commune. Research Lab., Southeast University (No.2015D08).

\end{document}